\documentclass{article}

\usepackage[preprint]{corl_2026} 

\usepackage{amsmath}
\usepackage{amssymb}
\usepackage{mathtools}
\usepackage{graphicx}
\usepackage{booktabs}
\usepackage{multirow}
\usepackage{caption}
\usepackage{array}
\usepackage{nicefrac}       
\usepackage{microtype}      
\usepackage{makecell}
\usepackage{algorithm}
\usepackage{algpseudocode}

\newcommand{\best}[1]{\textbf{#1}}
\newcommand{\second}[1]{\underline{#1}}

\usepackage[normalem]{ulem} 

\title{Anchor3R: Streaming 3D Reconstruction with Transient Anchors for Long-Horizon Visual Mapping}

\author{
\begin{tabular}{ccccc}
Peilin Tao\textsuperscript{1,2,3} &
Chong Cheng\textsuperscript{3,4} &
Yuansen Du\textsuperscript{3,\ensuremath{\ddagger}} &
Caiwei Song\textsuperscript{3} &
Zhengqing Chen\textsuperscript{3}
\end{tabular}
\\
\begin{tabular}{cccccc}
Xiaoyang Guo\textsuperscript{3} &
Wei Yin\textsuperscript{3} &
Weiqiang Ren\textsuperscript{3} &
Qian Zhang\textsuperscript{3} &
Hainan Cui\textsuperscript{1,2,\ensuremath{\dagger}} &
Shuhan Shen\textsuperscript{1,2,\ensuremath{\dagger}}
\end{tabular}
\\[0.6em]
\textsuperscript{1}Institute of Automation, Chinese Academy of Sciences \quad \\
\textsuperscript{2}School of Artificial Intelligence, University of Chinese Academy of Sciences \quad \\
\textsuperscript{3}Horizon Robotics \quad
\textsuperscript{4}HKUST(GZ)
\\[0.4em]
\textsuperscript{\ensuremath{\dagger}} Corresponding author \quad
\textsuperscript{\ensuremath{\ddagger}} Project lead
}

\begin{document}
\maketitle

\vspace{-0.5cm}
\begin{center}
  \includegraphics[trim=20bp 65bp 20bp 20bp, clip, width=\textwidth]{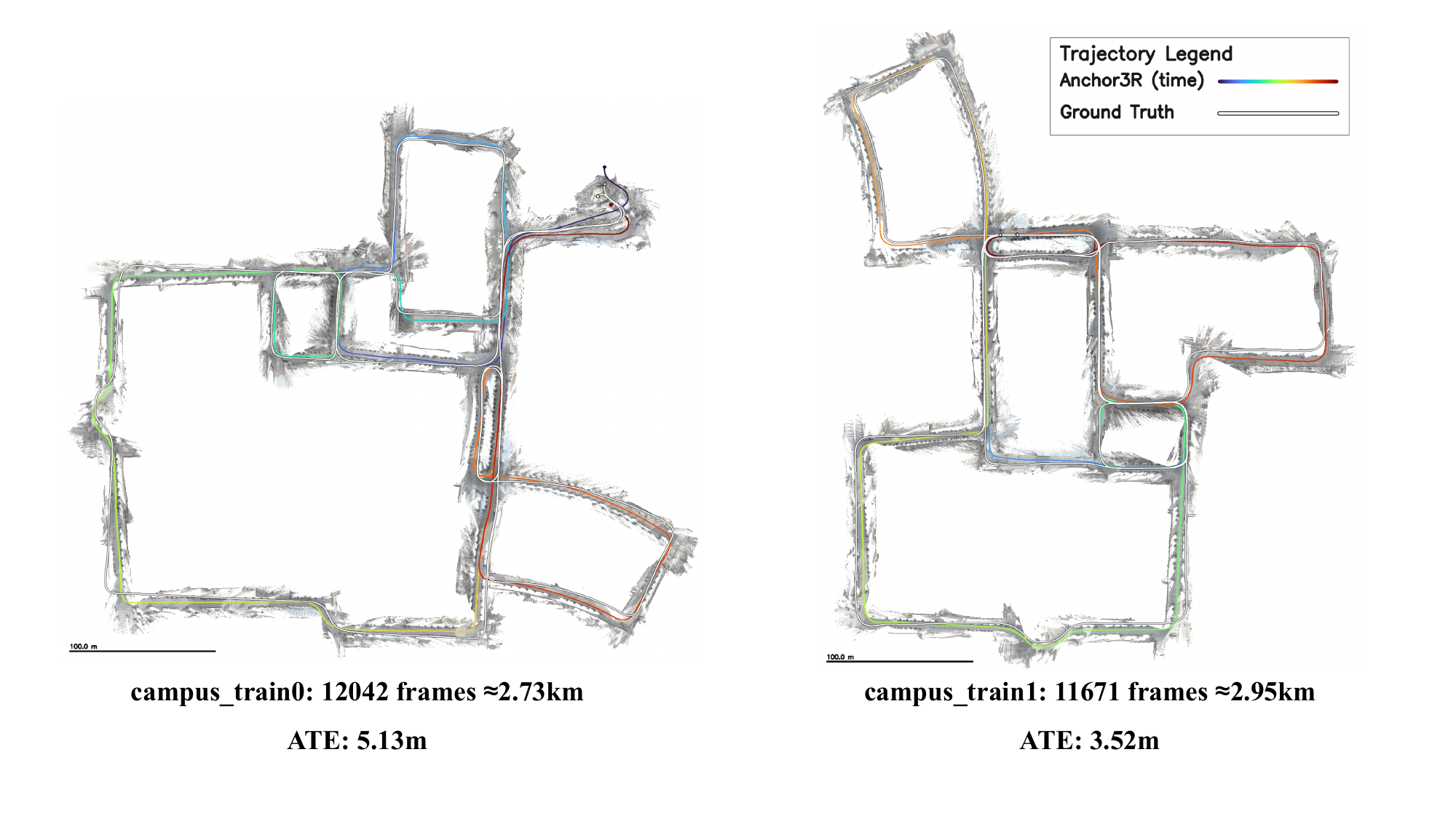}

  {
  \captionsetup{type=figure,hypcap=false}
\caption{
\textbf{Anchor3R results on campus-scale sequences.}
We visualize our reconstruction on campus\_train0 and campus\_train1 of the VBR dataset, two real-world sequences with \textbf{12,042} and \textbf{11,671} frames.
Predicted trajectories are color-coded and ground truth is shown as white curves.
}
  \label{fig:campus-loop}
  }
\end{center}
\vspace{-0.3cm}
\begin{abstract}
Long-horizon online visual mapping requires continuous camera-motion and scene-geometry estimation under bounded computation. Recent feed-forward 3D reconstruction models provide strong geometric priors, but streaming variants often predict poses in a fixed or historically maintained coordinate system, leading to train--test mismatch, early-anchor attention bias, and accumulated drift. We propose \emph{Anchor3R}, a current-centric streaming 3D reconstruction framework that predicts window-relative poses and local geometry in the current-frame coordinate system. Overlapping predictions form a dense relative-pose graph, supporting online pose updates and loop-aware motion averaging for global reconstruction. Experiments on indoor, outdoor, driving, and RGB-D benchmarks demonstrate improved long-horizon pose accuracy and dense reconstruction quality over existing streaming baselines. Despite being trained only on 48-frame sequences, Anchor3R directly generalizes to streams exceeding 10,000 frames while maintaining bounded GPU memory during online inference.
Code is available at \url{https://github.com/polar-explorer/Anchor3R}.
\end{abstract}


\section{Introduction}

Although recent streaming variants improve scalability through recurrent scene
memories~\citep{wang2025continuous,chen2025ttt3r,dong2026memix}, causal
Transformer caches~\citep{lan2025stream3r,zhuo2025streaming,cheng2026longstream},
or compact camera/state pools~\citep{li2025wint3r,liu2026mem3r}, most of them
still formulate streaming reconstruction as sequential state prediction in a
fixed or historically maintained gauge. This forces the model to propagate a
long-lived coordinate system through hidden states, entangling local geometric
reasoning with historical gauge maintenance. Such coupling can lead to
attention sinks around early anchors~\citep{cheng2026longstream} and causes
uncertainty and scale errors to accumulate over long sequences. Moreover, the
accumulated outputs form a trajectory rather than a graph of independent visual
measurements, making global error redistribution difficult. This raises a key
question: \textit{how should feed-forward 3D models expose their predictions
for large-scale streaming reconstruction?} We argue that streaming mapping
should decouple local measurement from global gauge alignment.

We propose \emph{Anchor3R}, a current-centric streaming 3D reconstruction
framework for long-horizon visual mapping. Instead of maintaining a persistent
global gauge, Anchor3R uses the current frame as a transient anchor and directly
predicts window-relative pose measurements
$\{\hat{\mathbf{T}}_{i\leftarrow t}\}_{i\in\mathcal{W}_t}$. This converts
streaming reconstruction from fixed-gauge state regression into repeated local
measurement prediction, allowing the network to focus on short-range relative
geometry reasoning without propagating long-range gauge drift through attention or
memory. As the window advances, overlapping predictions constrain each frame
from multiple anchors and naturally form a dense relative-pose graph, which
supports online pose updates and loop-aware motion averaging.

To realize this formulation, Anchor3R introduces a pose-query-based streaming
Transformer that separates frame-level visual evidence from window-level pose
reasoning. Decoupled frame attention computes image-conditioned states that can
be reused across windows, while current-centric window attention instantiates
pose-query tokens for the active window and reasons under the current-frame
gauge. We cache only image-side key/value states and discard pose-query states
after each local prediction. This image-only cache stores gauge-agnostic visual
and correspondence cues, while preventing pose tokens tied to one transient
gauge from being propagated into later windows. Therefore, Anchor3R achieves
bounded GPU-memory streaming inference without explicit cache refresh.

Finally, Anchor3R accumulates window-relative pose measurements into an explicit motion graph. In the online mode, the current pose is estimated from multiple measurements connected to previously recovered frames. 
For offline refinement, retrieved loop-closure keyframes are reinserted as transient anchors to generate long-range relative-pose edges, and motion averaging redistributes drift over the graph. 
The recovered global poses then align local point maps into a coherent reconstruction. 
Thus, the offline refinement is a direct consequence of the proposed prediction interface: Anchor3R operates on a dense relative-pose graph rather than a single predicted trajectory.

Our contributions are threefold:
1) We formulate streaming feed-forward reconstruction as current-centric
dense relative-pose prediction, replacing persistent global-gauge
regression with window-relative predictions anchored at the current frame.
2) We design an image-only cached pose-query Transformer that separates
reusable frame evidence from transient current-gauge pose reasoning, enabling
bounded GPU-memory streaming inference without propagating stale pose states.
3) We accumulate dense relative-pose measurements into a motion graph that
supports online pose updates and loop-aware motion averaging, enabling global
drift redistribution from overlapping local predictions.
\section{Related Work}

\paragraph{Classical SfM and SLAM.}
Classical SfM and SLAM systems remain strong baselines for visual mapping because they explicitly enforce multiview geometric constraints. Incremental SfM~\citep{schonberger2016structure} estimates pairwise relative poses from image correspondences and sequentially registers cameras using Perspective-$n$-Point estimation~\citep{ding2023revisiting}, triangulation~\citep{hartley2003multiple}, and bundle adjustment~\citep{triggs1999bundle,ren2022megba}. SLAM systems~\citep{mur2015orb,mur2017orb} further emphasize online tracking, local mapping, loop closure, and map refinement. Global SfM~\citep{pan2024global} instead estimates camera motions jointly from pairwise relative poses via rotation and translation averaging~\citep{zhu2018very,chatterjee2017robust,ozyesil2015robust}, followed by triangulation and refinement. Such local-to-global decomposition is attractive for long-horizon mapping as it separates relative measurement estimation from global alignment. Anchor3R follows the same principle, but replaces correspondence-based relative pose estimation with dense relative-pose prediction from a streaming feed-forward model.

\paragraph{Offline Feed-forward Reconstruction.}
Recent feed-forward visual geometry models aim to recover camera motion and dense structure directly from RGB inputs. DUSt3R~\citep{wang2024dust3r} introduced end-to-end two-view pointmap prediction in a shared pairwise coordinate frame. VGGT~\citep{wang2025vggt} further unified camera, depth, point map, and track prediction within a Transformer, while $\pi^3$~\citep{wang2025pi} removed fixed-reference bias through a permutation-equivariant formulation. More recent models extend this paradigm to broader settings and larger scales, including Depth Anything 3~\citep{lin2025depth}, MapAnything~\citep{keetha2025mapanything}, and VGG-T$^3$~\citep{elflein2026vgg}. Several works further improve scalability by processing long sequences or image collections through chunks, compact states, or test-time adaptation. VGGT-Long~\citep{deng2025vggt} uses chunk-wise reconstruction, overlap-based alignment, and loop closure; ZipMap~\citep{jin2026zipmap} compresses an image collection into a compact hidden scene state with test-time training layers; LoGeR~\citep{zhang2026loger} combines hybrid memory with chunked long-video reconstruction; and Scal3R~\citep{xie2026scal3r} introduces test-time-adapted global context for large-scale reconstruction. These methods provide powerful geometric priors and improve long-context scalability, but they are mainly designed for offline or chunk-based processing over fixed inputs. In contrast, Anchor3R targets streaming visual mapping, where the model must output local pose and geometry updates as frames arrive while still supporting global error redistribution over the accumulated motion graph.

\paragraph{Streaming Feed-forward Reconstruction.}
Native streaming reconstruction processes frames incrementally and must maintain useful historical information under bounded or partially bounded memory. CUT3R~\citep{wang2025continuous} uses persistent latent scene memory to support continuous prediction, while Point3R~\citep{wu2025point3r} introduces explicit spatial pointer memory anchored to reconstructed 3D structure. STream3R~\citep{lan2025stream3r} formulates sequential reconstruction with a causal Transformer, and WinT3R~\citep{li2025wint3r} combines sliding-window processing with a compact camera token pool. LongStream~\citep{cheng2026longstream} identifies first-frame anchoring as a major source of long-sequence degradation and predicts keyframe-relative poses to reduce gauge coupling. Recent methods such as TTT3R~\citep{chen2025ttt3r} and Mem3R~\citep{liu2026mem3r} further explore inference-time state updating and hybrid memory mechanisms. Despite these advances, many streaming methods still rely on the backbone memory or cache to implicitly maintain a coordinate gauge over time. As a result, local geometric prediction, historical state propagation, and long-range consistency become tightly coupled, making it difficult for later observations to correct accumulated drift. In contrast, Anchor3R decouples scalable bounded-window prediction from global consistency recovery: the network predicts current-centric window-relative poses as local measurements, and motion averaging integrates them into a globally consistent trajectory.

\section{Method}
\label{sec:method}

Anchor3R adopts a local-to-global design for long-horizon streaming reconstruction.
It uses the current frame as a transient anchor to predict relative-pose measurements within an active window, while local pointmaps are predicted in the current-frame coordinate system.
Across overlapping windows, these measurements form a dense relative-pose graph, which supports online pose updates and loop-aware motion averaging.
Thus, neural prediction focuses on short-range relative geometry, while graph optimization handles long-range gauge alignment and drift redistribution.

\subsection{Current-Centric Formulation} Given an image stream $\mathcal{I}=\{I_1,\dots,I_T\}$, Anchor3R processes the sequence with an active sliding window $\mathcal{W}_t=\{k,\dots,t\}$ at time $t$, where $k=\max(1,t-W+1)$. For each frame $i\in\mathcal{W}_t$, the network predicts a relative pose $\hat{\mathbf{T}}_{i\leftarrow t}\in\mathrm{SE}(3)$ from the current frame $I_t$ to frame $I_i$. Given global camera poses $\mathbf{T}_i\in\mathrm{SE}(3)$, the training target is \begin{equation} \mathbf{T}_{i\leftarrow t}=\mathbf{T}_i\mathbf{T}_t^{-1}, \qquad \mathbf{T}_{t\leftarrow t}=\mathbf{I}. \label{eq:current_centric_target} \end{equation} This target depends only on the relative geometry within the active window, rather than on an increasingly distant global origin. Therefore, the network can focus on local visual overlap, correspondence reasoning, and short-range geometric consistency, without propagating a long-lived gauge through hidden states or attention caches. In parallel, Anchor3R predicts a current-frame local pointmap $\hat{\mathbf{X}}_t^{\mathrm{local}}$. At time $t$, the prediction produces a set of graph edges $\mathcal{E}_t=\{(i,t,\hat{\mathbf{T}}_{i\leftarrow t})\mid i\in\mathcal{W}_t,\, i<t\}$. As the window slides, repeated observations from later anchors add redundant edges for the same frames, turning local window predictions into a dense relative-pose measurement graph. The recovered global poses then transform local pointmaps into a coherent reconstruction.

\begin{figure}
  \centering
  \includegraphics[trim=20bp 15bp 20bp 15bp, clip, width=0.95\textwidth]{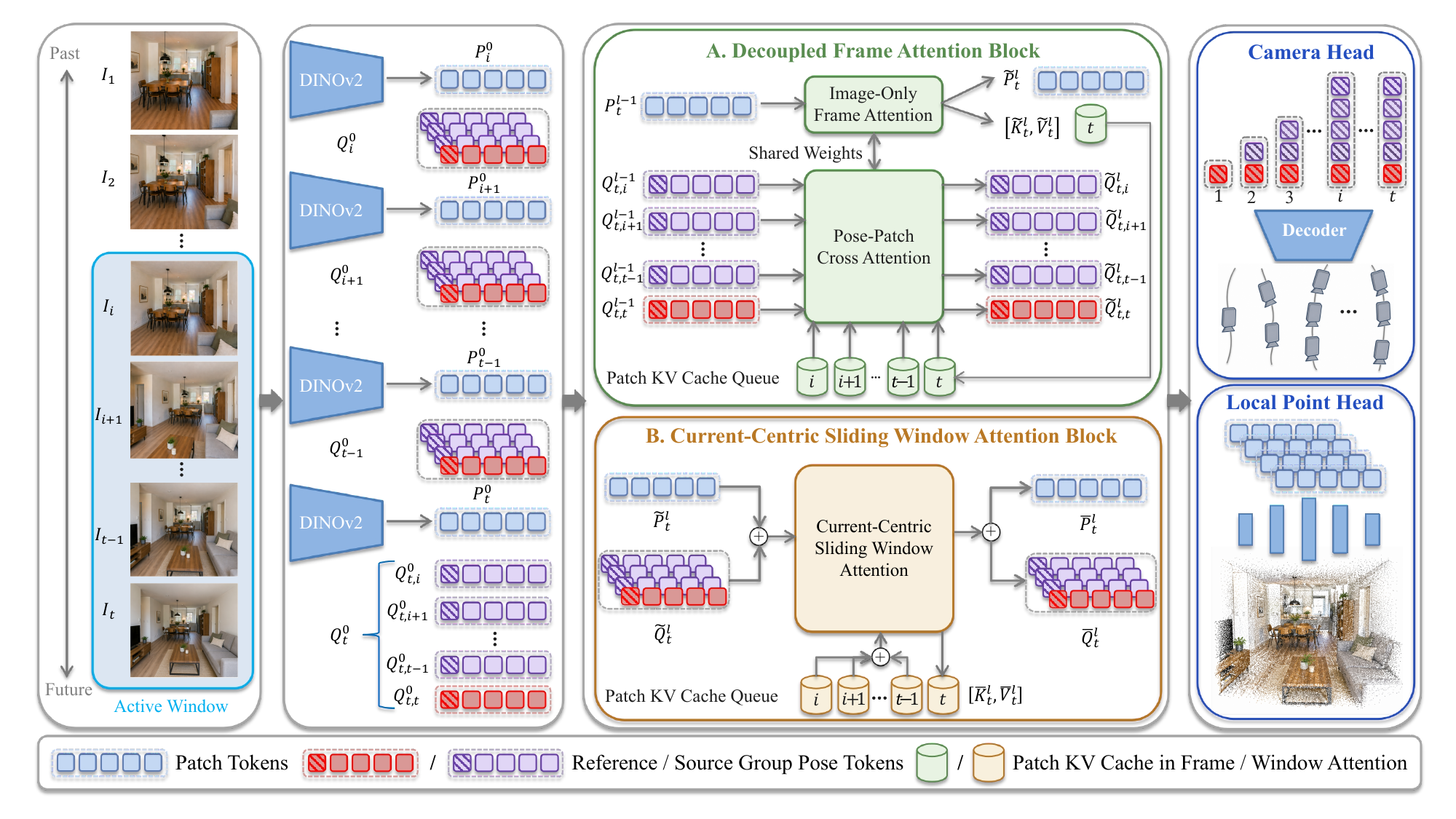}
  \caption{
  \textbf{Overview of Anchor3R.}
  Given the current frame $I_t$, Anchor3R extracts DINOv2 patch tokens and instantiates grouped pose-query tokens for frames in $\mathcal{W}_t$, using $I_t$ as the local reference. A sliding-window pose-query Transformer alternates between decoupled frame attention and current-centric window attention. The camera head decodes window-relative poses, while the point head predicts the current-frame pointmap.
  }
\vspace{-0.3cm}
\end{figure}

\subsection{Sliding-Window Pose-Query Transformer}

For each incoming frame $I_t$, a DINOv2 encoder extracts patch tokens
$\mathbf{P}_t^0\in\mathbb{R}^{N\times C}$, while historical frames provide
cached image-conditioned states.
For each frame $i\in\mathcal{W}_t$, we instantiate a pose-query group
$\mathbf{Q}_{t,i}^0\in\mathbb{R}^{(1+X)\times C}$, consisting of one pose token
and $X$ register tokens.
The full query set is
$\mathbf{Q}_t^0=\operatorname{Concat}_{i\in\mathcal{W}_t}\mathbf{Q}_{t,i}^0$,
where the current frame uses a learnable reference template
$\mathbf{Q}^{\mathrm{ref}}$ and source frames use $\mathbf{Q}^{\mathrm{src}}$.
This asymmetric initialization defines the current frame as the local reference
and lets source queries aggregate motion evidence relative to it.

At layer $l$, the \textbf{decoupled frame attention} block updates only the
current image tokens,
$(\tilde{\mathbf{P}}_t^{\,l},\tilde{\mathbf{K}}_t^{\,l},\tilde{\mathbf{V}}_t^{\,l})
=\mathcal{F}^{\,l}(\mathbf{P}_t^{\,l-1})$,
and caches the image-side keys and values.
Each pose-query group then reads from its corresponding frame representation,
$\tilde{\mathbf{Q}}_{t,i}^{\,l}
=\mathcal{F}^{\,l}(\mathbf{Q}_{t,i}^{\,l-1};
\tilde{\mathbf{K}}_i^{\,l},\tilde{\mathbf{V}}_i^{\,l})$,
where historical image states are reused from the cache.
Thus, frame-level visual evidence is computed once and remains independent of
future pose-query instantiations.
The \textbf{current-centric window attention} block jointly updates the current
patch tokens and all pose-query groups:
$(\mathbf{P}_t^{\,l},\{\mathbf{Q}_{t,i}^{\,l}\}_{i\in\mathcal{W}_t},
\bar{\mathbf{K}}_t^{\,l},\bar{\mathbf{V}}_t^{\,l})=
\mathcal{W}^{\,l}(\operatorname{Concat}(\tilde{\mathbf{P}}_t^{\,l},
\{\tilde{\mathbf{Q}}_{t,i}^{\,l}\}_{i\in\mathcal{W}_t});
\bar{\mathbf{K}}_{\mathcal{W}_t}^{\,l},\bar{\mathbf{V}}_{\mathcal{W}_t}^{\,l})$,
where $\bar{\mathbf{K}}_{\mathcal{W}_t}^{\,l}$ and
$\bar{\mathbf{V}}_{\mathcal{W}_t}^{\,l}$ concatenate cached image-side states
from frames in $\mathcal{W}_t\setminus\{t\}$.
This design separates reusable visual evidence from transient pose reasoning:
pose-query tokens are tied to a current-centric gauge, so caching them across
windows would mix stale coordinate hypotheses from different anchors.
In contrast, image-conditioned patch states store gauge-agnostic appearance,
geometry, and matching cues.
By caching only image-side states, Anchor3R keeps correspondence information in
patch tokens while rebuilding gauge-dependent pose reasoning for each window.



\subsection{Prediction Heads}
After the final layer, Anchor3R predicts window-relative poses and current-frame local geometry. For each query group $\mathbf{Q}_{t,i}^{\,L}$, its first token is used as the pose token, and the camera head jointly decodes $\{\hat{\mathbf{T}}_{i\leftarrow t}\}_{i\in\mathcal{W}_t}=\mathcal{H}_{\mathrm{cam}}(\operatorname{Concat}_{i\in\mathcal{W}_t}\mathbf{Q}_{t,i}^{\,L}[1])$. Joint decoding is used because all relative poses in the active window share the same current-frame reference.
In parallel, a DPT-style point head predicts $\big[\hat{\mathbf{X}}_t^{\mathrm{local}},\hat{\mathbf{\Sigma}}_t\big]=\mathcal{H}_{\mathrm{pt}}(\{\mathbf{P}_t^{\,l}\}_{l\in\mathcal{L}})\in\mathbb{R}^{H_0\times W_0\times 4}$ from selected multi-level patch features, where $\hat{\mathbf{X}}_t^{\mathrm{local}}$ is the current-frame pointmap and $\hat{\mathbf{\Sigma}}_t$ is the confidence map in the local camera coordinate system.

\subsection{Motion Graph and Pose Recovery}

The predicted relative poses define a motion graph $\mathcal{G}=(\mathcal{V},\mathcal{E})$, where vertices are frames and edges store window-relative measurements. Let the global pose of frame $i$ be represented by rotation $\mathbf{R}_i$ and camera center $\mathbf{c}_i$. For an edge $(i,t)$, the network predicts $(\hat{\mathbf{R}}_{i\leftarrow t},\hat{\boldsymbol{t}}_{i\leftarrow t})$, with $\hat{\mathbf{R}}_{i\leftarrow t}\approx \mathbf{R}_i\mathbf{R}_t^\top$ and $\hat{\boldsymbol{v}}_{i,t}=\hat{\mathbf{R}}_i^\top\hat{\boldsymbol{t}}_{i\leftarrow t}\approx \mathbf{c}_t-\mathbf{c}_i$. Instead of composing poses along a spanning tree, we recover global poses by averaging over the full graph, which exploits redundant local predictions and redistributes errors across the trajectory. We fix the first-frame gauge by setting $\hat{\mathbf{R}}_1=\mathbf{I}$ and $\hat{\mathbf{c}}_1=\mathbf{0}$.

For online inference at time $t$, Anchor3R estimates the current global pose from the $|\mathcal{W}_t|-1$ newly predicted relative poses $\{\hat{\mathbf{T}}_{i\leftarrow t}\}_{i\in\mathcal{W}_t\setminus\{t\}}$ and the previously estimated global poses $\{(\hat{\mathbf{R}}_i,\hat{\mathbf{c}}_i)\}_{i\in\mathcal{W}_t\setminus\{t\}}$. Each historical frame induces a candidate current rotation and center; we take the Lie-algebra median of the candidate rotations for $\hat{\mathbf{R}}_t$ and the coordinate-wise median of candidate centers $\hat{\mathbf{c}}_i+\hat{\boldsymbol{v}}_{i,t}$ for $\hat{\mathbf{c}}_t$. For offline refinement, we jointly optimize all rotations and centers:
\begin{equation}
\min_{\{\mathbf{R}_i\}}\sum_{(i,t)\in\mathcal{E}}
\rho_R\!\left(\left\|\operatorname{Log}\!\big(\hat{\mathbf{R}}_{i\leftarrow t}^{\top}\mathbf{R}_i\mathbf{R}_t^\top\big)\right\|_2\right),
\qquad
\min_{\{\mathbf{c}_i\}}\sum_{(i,t)\in\mathcal{E}}
\left\|\mathbf{c}_i-\mathbf{c}_t+\hat{\boldsymbol{v}}_{i,t}\right\|_1 .
\label{eq:motion_averaging}
\end{equation}
The rotation objective is solved by IRLS in the Lie algebra, and the translation objective is solved by ADMM~\citep{boyd2011distributed}. Historical or loop-closure keyframes can be reinserted into the active window to add long-range edges to $\mathcal{G}$, enabling global error redistribution beyond local streaming updates.

\subsection{Training}
\paragraph{Training objectives.}

Following VGGT~\citep{wang2025vggt}, we train Anchor3R with a multi-task objective
$\mathcal{L}=\lambda_{\mathrm{cam}}\mathcal{L}_{\mathrm{cam}}+\mathcal{L}_{\mathrm{pmap}}$,
where $\lambda_{\mathrm{cam}}$ balances the camera and point-map losses. The camera loss supervises the relative camera parameters predicted within each active window:
$\mathcal{L}_{\mathrm{cam}}=\sum_t\sum_{i\in\mathcal{W}_t}
\big(\|q_{i\leftarrow t}-\hat{q}_{i\leftarrow t}\|_1+
\|\mathbf{t}_{i\leftarrow t}-s^{*}\hat{\mathbf{t}}_{i\leftarrow t}\|_1\big)$,
where $q_{i\leftarrow t}$ and $\mathbf{t}_{i\leftarrow t}$ denote the ground-truth relative rotation and translation, respectively. The local point-map loss combines confidence-weighted reconstruction with gradient regularization:
$\mathcal{L}_{\mathrm{pmap}}=\sum_t
\|\hat{\mathbf{\Sigma}}_t\odot(\mathbf{X}_t^{\mathrm{local}}-s^{*}\hat{\mathbf{X}}_t^{\mathrm{local}})\|
+\|\hat{\mathbf{\Sigma}}_t\odot(\nabla\mathbf{X}_t^{\mathrm{local}}-\nabla(s^{*}\hat{\mathbf{X}}_t^{\mathrm{local}}))\|
-\alpha\log\hat{\mathbf{\Sigma}}_t$,
where $\hat{\mathbf{\Sigma}}_t$ is the predicted confidence map. 
The scale factor is estimated following $\pi^3$~\citep{wang2025pi} as
$s^{*}=\operatorname*{arg\,min}_{s}\sum_t
\|\mathbf{X}_t^{\mathrm{local}}-s\hat{\mathbf{X}}_t^{\mathrm{local}}\|_1$.
We apply the same $s^{*}$ to both relative camera translations and local point maps, encouraging the predicted cameras and geometry to remain scale-consistent within each active window and across overlapping windows.

\paragraph{Training data.}
We train Anchor3R on a mixture of real and synthetic datasets, including WildRGB~\citep{xia2024rgbd}, ScanNet~\citep{dai2017scannet}, HyperSim~\citep{roberts2021hypersim}, Mapillary~\citep{antequera2020mapillary}, Replica~\citep{straub2019replica}, Mapfree~\citep{arnold2022map}, TartanAir~\citep{wang2020tartanair}, MVS-Synth~\citep{huang2018deepmvs}, Virtual KITTI~\citep{cabon2020vkitti2}, Aria Synthetic Environments~\citep{pan2023aria}, Spring~\citep{mehl2023spring}, Waymo Open~\citep{sun2020scalability}, BlendedMVS~\citep{yao2020blendedmvs}, Co3Dv2~\citep{reizenstein2021common}, MegaDepth~\citep{li2018megadepth} and DL3DV~\citep{ling2024dl3dv}. For unordered data, we construct overlapping sequences with pose-guided sampling; for video data, we use random interval sampling and block shuffling to increase diversity while preserving local continuity.

\paragraph{Implementation details.}
Anchor3R is initialized from VGGT and retains its 24-layer backbone with
alternating attention block, resulting in approximately 1.2B parameters.
During training, we unroll each sample over 48 frames, while using a fixed
active window of $W=10$ frames for local prediction.
Each pose-query group contains one pose token and 31 register tokens.
The model is optimized with AdamW using cosine learning-rate decay, a peak
learning rate of $1\times10^{-4}$, and 2k warm-up steps. Images are resized
such that their longer side is at most 518 pixels, with aspect-ratio jittering
applied for data augmentation. Training is performed for 80k iterations on
32 NVIDIA A800 GPUs and takes approximately 12 days.

\section{Experiments}

We evaluate Anchor3R for long-horizon robot visual mapping through one-pass streaming pose estimation, loop-aware graph refinement, dense reconstruction, and controlled analyses of its current-centric formulation and streaming scalability.
For camera pose estimation, we use KITTI Odometry~\citep{Geiger2012CVPR},
VBR~\citep{brizi2024vbr}, TUM RGB-D~\citep{sturm2012benchmark},
Oxford Spires~\citep{tao2025spires}, and Waymo~\citep{sun2020scalability},
covering driving, handheld/vehicle mapping, indoor RGB-D, and mobile mapping
trajectories.
All main evaluation sequences are excluded from training; for Waymo, we use a
held-out subset of the official training split.
For dense reconstruction, we evaluate on 7Scenes~\citep{shotton2013scene}
and TUM RGB-D.
We further conduct controlled ablations on Virtual KITTI~\citep{cabon2020vkitti2}
with a ViT-Small backbone.

\begin{table*}[t]
\centering
\caption{\textbf{Quantitative comparison on KITTI~\citep{Geiger2012CVPR} in terms of ATE.}
The upper block lists optimization-based baselines, and the lower block reports streaming feed-forward methods.
Failed cases are marked by $*$ and excluded from the average.
Anchor3R achieves the best average accuracy.}
\label{tab:kitti_ate}
\resizebox{\textwidth}{!}{
\begin{tabular}{l|ccccccccccc|c}
\toprule
\multirow{2}{*}{\textbf{Methods}}
& \multicolumn{11}{c|}{\textbf{KITTI~\citep{Geiger2012CVPR} (ATE $\downarrow$)}} 
& \multirow{2}{*}{\textbf{Avg.}} \\
\cmidrule(lr){2-12}
& 00 & 01 & 02 & 03 & 04 & 05 & 06 & 07 & 08 & 09 & 10 & \\
& {\scriptsize 4541$\times$, 3.7\,km}
& {\scriptsize 1101$\times$, 2.5\,km}
& {\scriptsize 4661$\times$, 5.1\,km}
& {\scriptsize 801$\times$, 0.6\,km}
& {\scriptsize 271$\times$, 0.4\,km}
& {\scriptsize 2761$\times$, 2.2\,km}
& {\scriptsize 1101$\times$, 1.2\,km}
& {\scriptsize 1101$\times$, 0.7\,km}
& {\scriptsize 4071$\times$, 3.2\,km}
& {\scriptsize 1591$\times$, 1.7\,km}
& {\scriptsize 1201$\times$, 0.9\,km}
& \\
\midrule
FastVGGT
& * & 705.39 & * & 62.38 & 10.27 & 157.74 & 124.43 & 69.27 & * & 190.10 & 194.75 & 189.29 \\
MASt3R-SLAM
& * & 530.37 & * & 18.87 & 88.98 & 159.430 & 92.00 & * & * & * & * & 177.93 \\
VGGT-SLAM
& * & 607.16 & * & 169.83 & 13.12 & * & * & * & * & * & * & 263.37 \\
VGGT-Long
& \best{8.64} & \best{21.20} & \best{52.72} & 8.78 & 4.20 & \second{9.88} & \best{4.67} & \best{2.66} & 72.98 & \best{31.84} & 27.71 & \second{25.94} \\
\midrule
CUT3R
& 185.89 & 651.52 & 296.98 & 148.06 & 22.17 & 155.61 & 132.54 & 77.03 & 238.39 & 205.94 & 193.39 & 209.78 \\
TTT3R
& 190.93 & 546.84 & 218.77 & 105.28 & 11.62 & 153.12 & 132.94 & 70.95 & 180.57 & 211.01 & 133.00 & 177.73 \\
STream3R
& 190.98 & 681.95 & 301.40 & 158.25 & 102.73 & 159.85 & 135.03 & 90.37 & 261.15 & 216.31 & 207.49 & 227.77 \\
StreamVGGT
& 191.93 & 653.06 & 303.35 & 157.50 & 108.24 & 160.46 & 133.71 & 89.00 & 263.95 & 216.69 & 209.80 & 226.15 \\
LongStream
& 92.55
& \second{46.01}
& 134.70
& \best{3.81}
& \second{1.95}
& 84.69
& 23.12
& 14.93
& 62.07
& 85.61
& 21.48
& 51.90 \\
\midrule
\textbf{Anchor3R-Online}
& 44.18
& 48.43
& 149.61
& 4.39
& 1.99
& 62.63
& 12.12
& 10.39
& \second{52.12}
& 55.23
& \second{8.57}
& 40.89 \\
\textbf{Anchor3R-Offline}
& \second{19.68}
& 48.62
& \second{75.75}
& \second{4.22}
& \best{1.90}
& \best{7.56}
& \second{6.36}
& \second{7.63}
& \best{49.34}
& \second{49.02}
& \best{8.00}
& \best{25.03} \\
\bottomrule
\end{tabular}
}
\end{table*}

\begin{table*}[t]
\vspace{-8pt}
\centering
\caption{\textbf{Quantitative comparison} on VBR~\citep{brizi2024vbr}. We report ATE, where lower is better.}
\vspace{-5pt}
\label{tab:vbr_ate}
\setlength{\tabcolsep}{3.5pt}
\renewcommand{\arraystretch}{1.12}
\resizebox{\textwidth}{!}{
\begin{tabular}{l|ccccccc|c}
\toprule
\multirow{2}{*}{\textbf{Method}}
& \multicolumn{7}{c|}{\textbf{VBR~\citep{brizi2024vbr} ATE $\downarrow$}}
& \multirow{2}{*}{\textbf{Avg.}} \\
\cmidrule(lr){2-8}
& campus\_train0
& campus\_train1
& ciampino\_train1
& colosseo\_train0
& diag\_train0
& pincio\_train0
& spagna\_train0
& \\
& {\scriptsize 12042$\times$, 2.73\,km}
& {\scriptsize 11671$\times$, 2.95\,km}
& {\scriptsize 18846$\times$, 5.20\,km}
& {\scriptsize 8815$\times$, 1.45\,km}
& {\scriptsize 10021$\times$, 1.02\,km}
& {\scriptsize 11142$\times$, 1.27\,km}
& {\scriptsize 14141$\times$, 1.56\,km}
& \\
\midrule

VGGT-SLAM
& 93.51 & 71.74 & 124.10 & 101.00 & 33.64 & 66.42 & 57.00 & 78.20 \\

VGGT-Long
& 118.59 & 98.21 & 172.13 & \second{39.56} & 30.80 & 53.44 & 50.27 & 80.43 \\

Pi3-Chunk
& \second{78.50} & \second{65.77} & \second{111.72} & 77.09 & \second{23.81} & \second{41.99} & \second{44.76} & \second{63.38} \\

\midrule

InfiniteVGGT
& 123.65 & 100.00 & * & 83.91 & 31.58 & 70.73 & 56.25 & 91.60 \\

LongStream
& 100.57 & 105.55 & 131.78 & 72.52 & 32.35 & 43.47 & 59.31 & 77.93 \\

\midrule

\textbf{Anchor3R-Online}
& 86.63 & 82.16 & 168.25 & 61.43 & 29.38 & 51.34 & 54.34 & 76.21 \\

\textbf{Anchor3R-Offline}
& \best{5.13} & \best{3.52} & \best{78.64} & \best{17.25} & \best{5.35} & \best{15.56} & \best{11.76} & \best{19.60} \\

\bottomrule
\end{tabular}
}
\vspace{-12pt}
\end{table*}

\subsection{Camera Pose Estimation}

\paragraph{Evaluation protocol.}
We align each predicted trajectory to the ground truth with a similarity transformation and report Absolute Trajectory Error (ATE), where lower is better. Failed cases are marked by $*$ and excluded from the average. We compare against optimization-based systems, including FastVGGT~\citep{shen2025fastvggt}, MASt3R-SLAM~\citep{murai2025mast3r}, VGGT-SLAM~\citep{maggio2025vggt}, VGGT-Long~\citep{deng2025vggt} and Pi3-Chunk~\citep{wang2025pi}, as well as streaming feed-forward methods, including CUT3R~\citep{wang2025continuous}, TTT3R~\citep{chen2025ttt3r}, STream3R~\citep{lan2025stream3r}, StreamVGGT~\citep{zhuo2025streaming}, LongStream~\citep{cheng2026longstream}, and InfiniteVGGT~\citep{yuan2026infinitevggt}. 
We report two variants. Anchor3R-Online follows a strict one-pass streaming protocol: frames are processed in temporal order, poses are incrementally updated from current-centric window measurements, and no loop detection or loop-keyframe reinsertion is used. Anchor3R-Offline evaluates the graph-refinement capability enabled by our prediction interface. It augments the accumulated relative-pose graph with loop-closure measurements and performs motion averaging. Specifically, we select keyframes every 5 frames, retrieve the top-3 similar images using NetVLAD~\citep{arandjelovic2016netvlad}, and reinsert temporally separated matches as transient anchors to predict long-range relative-pose measurements.

\paragraph{Results.}

Table~\ref{tab:kitti_ate} reports results on KITTI.
Existing streaming feed-forward methods still show large errors on kilometer-scale trajectories, suggesting that recurrent states or causal caches alone are insufficient for stable long-horizon pose estimation.
Under the strict online protocol without loop closure, Anchor3R-Online reduces the average ATE from 51.90\,m for LongStream to 40.89\,m, demonstrating the benefit of current-centric local measurement prediction independently of offline refinement.
Anchor3R-Offline further reduces the error to 25.03\,m by optimizing the accumulated dense relative-pose graph.
Table~\ref{tab:vbr_ate} evaluates longer and more diverse real-world routes on VBR.
Anchor3R-Online remains competitive with prior streaming methods, while Anchor3R-Offline achieves the best average ATE and ranks first on all seven sequences among the methods in Table~\ref{tab:vbr_ate}.
The larger Online--Offline improvement on VBR is consistent with the availability of long-range revisits: most VBR trajectories contain loop closures, whereas only some KITTI sequences do.
This suggests that current-centric local measurements are useful both for strict online estimation and as reusable constraints for later global refinement.
Table~\ref{tab:tum_oxford_waymo} further demonstrates generalization across indoor RGB-D, mobile mapping, and driving scenarios.

\begin{table*}[t]
\centering

\begin{minipage}[t]{0.48\textwidth}
\centering
\captionof{table}{
\textbf{Quantitative comparison on TUM, Oxford Spires, and Waymo.}
Top: optimization-based methods; Bottom: streaming methods.
Anchor3R remains robust across short indoor trajectories and mobile mapping sequences.
}
\label{tab:tum_oxford_waymo}
\resizebox{\linewidth}{!}{
\begin{tabular}{lccc}
\toprule
\multirow{2}{*}{Methods}
& TUM~\citep{sturm2012benchmark}
& Oxford Spires~\citep{tao2025spires}
& Waymo~\citep{sun2020scalability} \\
\cmidrule(lr){2-4}
& ATE $\downarrow$
& ATE $\downarrow$
& ATE $\downarrow$ \\
\midrule
FastVGGT    & 0.418 & 36.577 & 1.281 \\
MASt3R-SLAM & 0.082 & 37.728 & 7.625 \\
VGGT-SLAM   & 0.123 & 31.003 & 7.431 \\
\midrule
CUT3R       & 0.542 & 32.440 & 9.396 \\
TTT3R       & 0.308 & 36.214 & 3.486 \\
STream3R    & 0.633 & 37.569 & 42.203 \\
StreamVGGT  & 0.627 & 37.255 & 45.101 \\
LongStream  & \textbf{0.076} & {19.815} & {0.737} \\
\midrule
\textbf{Anchor3R-Online}   & {0.091} & \textbf{17.661} & \textbf{0.425} \\
\bottomrule
\end{tabular}
}
\end{minipage}
\hfill
\begin{minipage}[t]{0.48\textwidth}
\centering
\captionof{table}{
\textbf{Quantitative comparison on 7Scenes and TUM.}
CD $\downarrow$ and F1@0.25 $\uparrow$ are adopted for evaluation.
Best numbers are in bold; second-best numbers are underlined.
}
\label{tab:geometry_7scenes_tum}
\resizebox{\linewidth}{!}{
\begin{tabular}{lcc|cc}
\toprule
\multirow{2}{*}{Methods}
& \multicolumn{2}{c|}{7Scenes}
& \multicolumn{2}{c}{TUM} \\
\cmidrule(lr){2-3}
\cmidrule(lr){4-5}
& CD $\downarrow$ & F1@0.25 $\uparrow$
& CD $\downarrow$ & F1@0.25 $\uparrow$ \\
\midrule
FastVGGT    & 6.373 & \textbf{0.710} & \underline{0.104} & {0.926} \\
MASt3R-SLAM & 5.987 & 0.691 & \textbf{0.057} & \textbf{0.954} \\
VGGT-SLAM   & 6.306 & {0.696} & 1.993 & 0.633 \\
\midrule
CUT3R       & 6.281 & 0.274 & 0.474 & 0.533 \\
TTT3R       & 6.231 & 0.260 & 0.249 & 0.792 \\
STream3R    & {6.353} & 0.479 & 1.126 & 0.444 \\
StreamVGGT  & 6.630 & {0.483} & 0.680 & 0.402 \\
LongStream  & \underline{2.260} & 0.641 & 0.225 & 0.673 \\
\midrule
\textbf{Anchor3R-Online}    & \textbf{1.848} & \underline{0.707} & 0.108 & \underline{0.933} \\
\bottomrule
\end{tabular}
}
\end{minipage}
\vspace{-10pt}
\end{table*}
\subsection{3D Reconstruction}

\paragraph{Evaluation protocol.}
We evaluate dense reconstruction on 7Scenes~\citep{shotton2013scene} and TUM RGB-D~\citep{sturm2012benchmark}. For each method, we reconstruct a point cloud from the predicted camera trajectory and depth or point-map outputs, align it to the ground truth with a similarity transformation, and report Chamfer Distance (CD) and F1 score at a threshold of 0.25.

\paragraph{Results.}
Table~\ref{tab:geometry_7scenes_tum} compares online reconstruction quality.
Anchor3R-Online achieves the best CD on 7Scenes and the second-best F1 score,
substantially improving over streaming baselines such as CUT3R~\citep{wang2025continuous},
TTT3R~\citep{chen2025ttt3r}, STream3R~\citep{lan2025stream3r}, and
StreamVGGT~\citep{zhuo2025streaming}. On TUM, MASt3R-SLAM~\citep{murai2025mast3r}
performs best due to SLAM-style geometric optimization on short indoor
trajectories, while Anchor3R-Online remains competitive with the second-best
F1 score. These results show that even without loop-aware offline refinement,
current-centric streaming pose updates provide sufficiently stable camera
estimates to align local pointmaps into coherent reconstructions.



\subsection{Ablation and Analysis}

We further conduct controlled analyses to study the current-centric formulation, cache design, active-window size, streaming scalability, and the effect of offline refinement. Unless otherwise specified, diagnostic ablations are conducted on Virtual KITTI~\citep{cabon2020vkitti2} with a ViT-Small backbone.

\begin{figure*}[t]
\centering
\begin{minipage}[t]{0.47\textwidth}
  \centering
  \includegraphics[trim=60bp 0bp 40bp 0bp, clip, width=\textwidth]{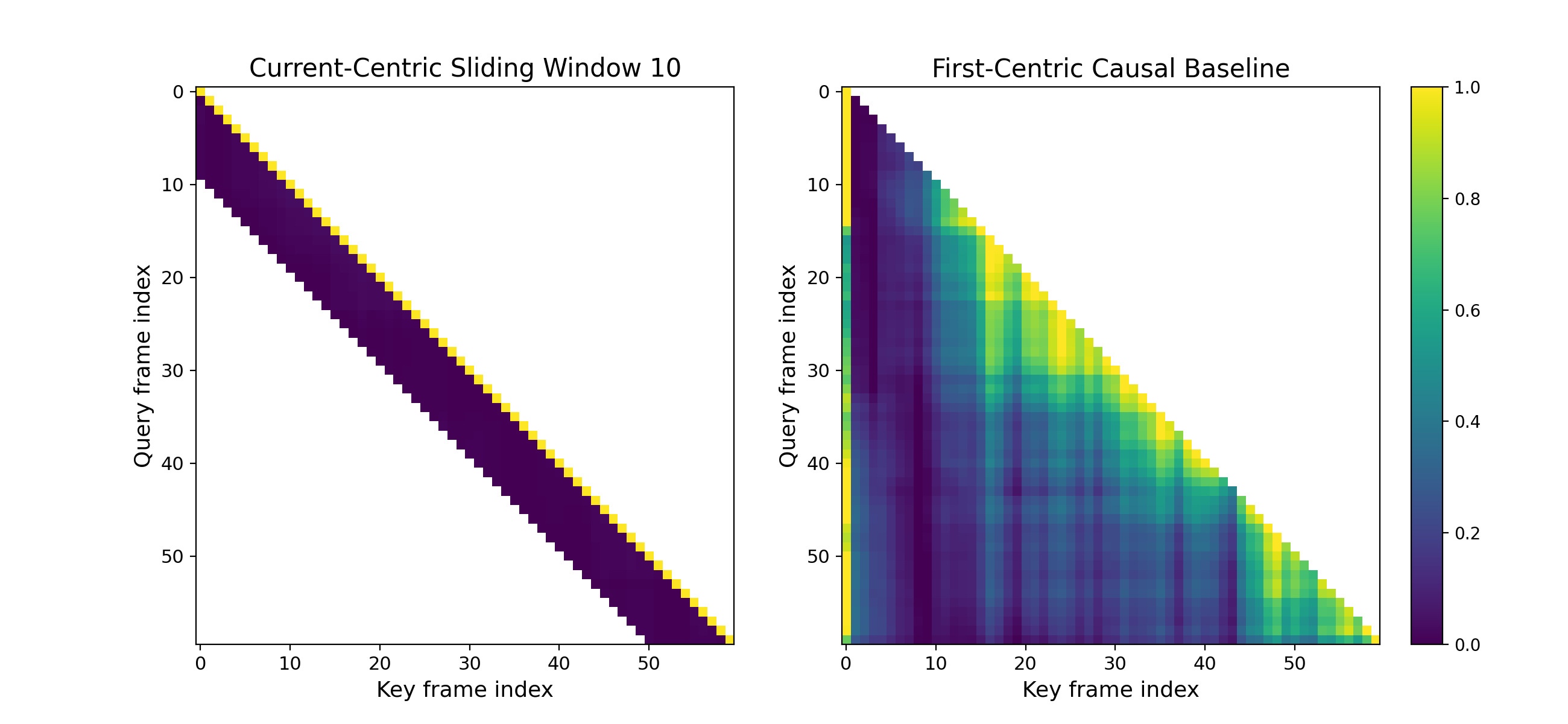}
  \captionof{figure}{
  \textbf{Attention-score comparison.}
  The first-centric baseline exhibits persistent attention to the first frame
throughout the sequence, while Anchor3R shifts its attention support forward
with the current-centric sliding window.
  }
  \label{fig:attention_score_ablation}
\end{minipage}
\hfill
\begin{minipage}[t]{0.51\textwidth}
  \centering
  \includegraphics[trim=5bp 0bp 5bp 0bp, clip, width=\textwidth]{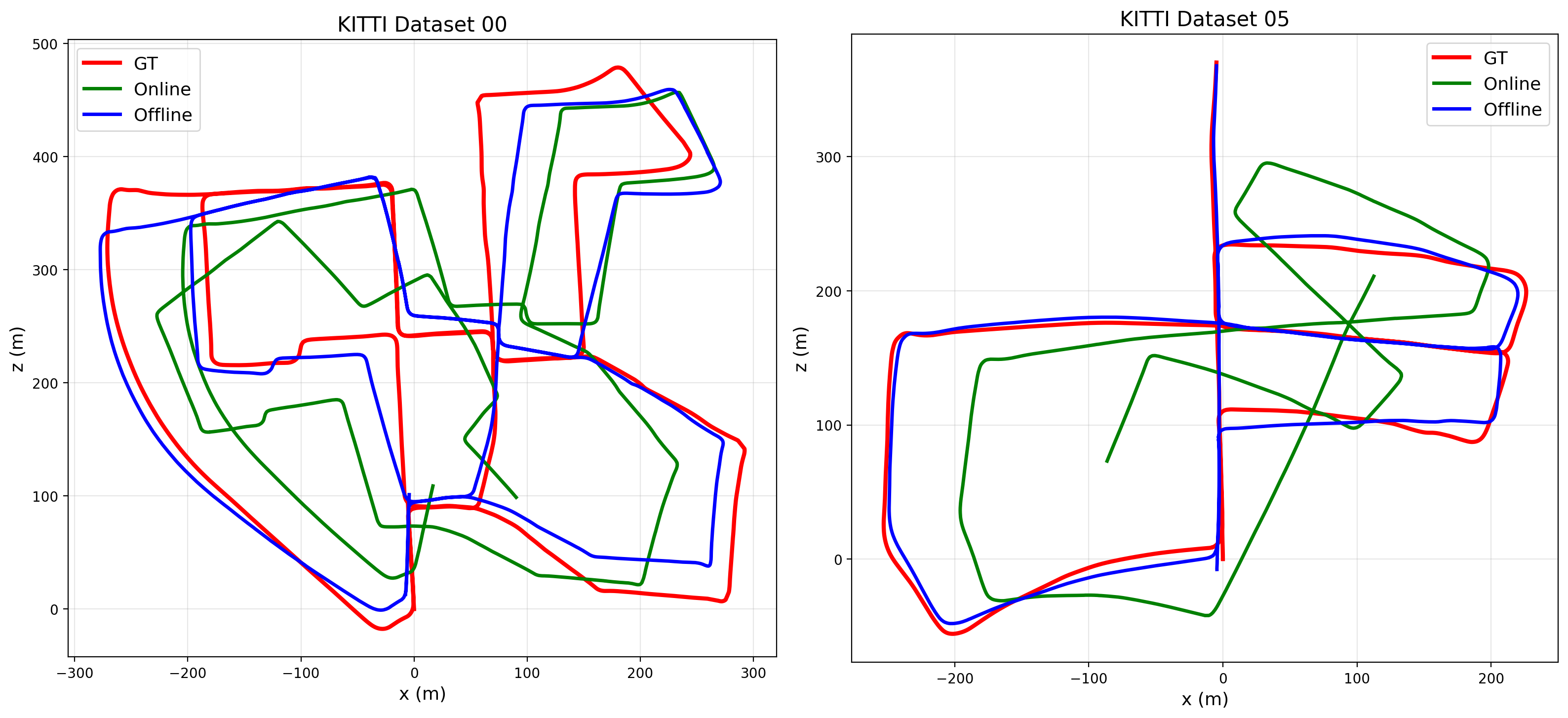}
  \captionof{figure}{
  \textbf{Qualitative comparison on KITTI.}
Loop-closure constraints make the offline trajectories of sequences 00 and 05 more globally consistent and closer to the ground truth.
  }
  \label{fig:kitti-loop}
\end{minipage}
\vspace{-6pt}
\end{figure*}

\paragraph{Current-centric prediction.}
A STream3R-like fixed-gauge streaming baseline~\citep{lan2025stream3r} implicitly uses the first frame as a persistent global anchor. As observed in LongStream~\citep{cheng2026longstream}, such first-frame anchoring can create an attention sink, especially when later frames become weakly overlapping with the initial view. 
Figure~\ref{fig:attention_score_ablation} visualizes attention over 60 vKITTI
frames. The first-centric baseline retains strong attention to the first frame
throughout the sequence, forming a persistent first-frame attention pattern
even as the query advances, while Anchor3R shifts its attention support forward
with the current-centric sliding window. This result supports our first
contribution: replacing persistent global-gauge regression with current-centric
local measurement prediction reduces the burden of long-range coordinate
maintenance inside the network.


\paragraph{Image-only key/value cache.}
Table~\ref{tab:vkitti_ate_rre_rte} compares two cache designs for current-centric sliding-window attention. The first variant caches both image tokens and pose-query tokens, while the second caches only image-conditioned states and re-instantiates pose queries for each current-centric window. Removing pose-query tokens from the cache improves ATE, RTE, and RRE on most scenes. This confirms that pose-query states are not generic reusable memory: they encode reasoning under a particular local gauge and can interfere with later predictions when propagated across windows. In contrast, image-conditioned states serve as reusable visual evidence and can be safely cached without explicit cache refresh. This ablation supports our second contribution: bounded GPU-memory streaming should cache image evidence while keeping pose reasoning transient and current-gauge-specific.

\begin{table*}[t]
\centering
\caption{
\textbf{Quantitative comparison on vKITTI.}
For each scene, we report ATE, relative rotation error (RRE), and relative translation error (RTE).
The number of images and trajectory length are shown under each scene name.
Best results are highlighted in bold.
}
\label{tab:vkitti_ate_rre_rte}
\resizebox{\textwidth}{!}{
\begin{tabular}{lccc ccc ccc ccc ccc|ccc}
\toprule
\multirow{3}{*}{Methods}
& \multicolumn{3}{c}{Scene 01}
& \multicolumn{3}{c}{Scene 02}
& \multicolumn{3}{c}{Scene 06}
& \multicolumn{3}{c}{Scene 18}
& \multicolumn{3}{c|}{Scene 20}
& \multicolumn{3}{c}{Avg.} \\
& \multicolumn{3}{c}{$447\times,\,332\,\mathrm{m}$}
& \multicolumn{3}{c}{$223\times,\,113\,\mathrm{m}$}
& \multicolumn{3}{c}{$270\times,\,51\,\mathrm{m}$}
& \multicolumn{3}{c}{$339\times,\,254\,\mathrm{m}$}
& \multicolumn{3}{c|}{$837\times,\,711\,\mathrm{m}$}
& \multicolumn{3}{c}{} \\
\cmidrule(lr){2-4}
\cmidrule(lr){5-7}
\cmidrule(lr){8-10}
\cmidrule(lr){11-13}
\cmidrule(lr){14-16}
\cmidrule(lr){17-19}
& ATE $\downarrow$ & RTE $\downarrow$ & RRE $\downarrow$
& ATE $\downarrow$ & RTE $\downarrow$ & RRE $\downarrow$
& ATE $\downarrow$ & RTE $\downarrow$ & RRE $\downarrow$
& ATE $\downarrow$ & RTE $\downarrow$ & RRE $\downarrow$
& ATE $\downarrow$ & RTE $\downarrow$ & RRE $\downarrow$
& ATE $\downarrow$ & RTE $\downarrow$ & RRE $\downarrow$ \\
\midrule
w pose cache
& 3.613 & 0.177 & 0.132
& 2.174 & 0.111 & 0.074
& \textbf{0.285} & \textbf{0.037} & \textbf{0.055}
& 3.955 & 0.195 & 0.087
& 7.558 & 0.211 & 0.112
& 3.517 & 0.146 & 0.092
 \\
w/o pose cache
& \textbf{3.113} & \textbf{0.101} & \textbf{0.127}
& \textbf{1.054} & \textbf{0.084} & \textbf{0.073}
& 0.424 & 0.042 & 0.059
& \textbf{1.410} & \textbf{0.118} & \textbf{0.080}
& \textbf{7.208} & \textbf{0.142} & \textbf{0.110}
& \textbf{2.642} & \textbf{0.097} & \textbf{0.090}
 \\
\bottomrule
\end{tabular}
}
\vspace{-12pt}
\end{table*}

\paragraph{Active window size.}
We vary $W\in \{5,10,15\}$ using the ViT-Small model on vKITTI. Increasing $W$ from 5 to 10 improves ATE/RRE/RTE from 3.23/0.12/0.15 to 2.64/0.09/0.10, while $W=15$ provides only marginal gains (2.55/0.08/0.10) at higher cost (3.2 vs.\ 2.7\,s/iter; 22.9 vs.\ 27.5 FPS). We therefore use $W=10$ as a balance between accuracy and efficiency.

\begin{figure}[t]
    \centering
    \includegraphics[
        width=\linewidth,
        trim=0 0 0 0,
        clip
    ]{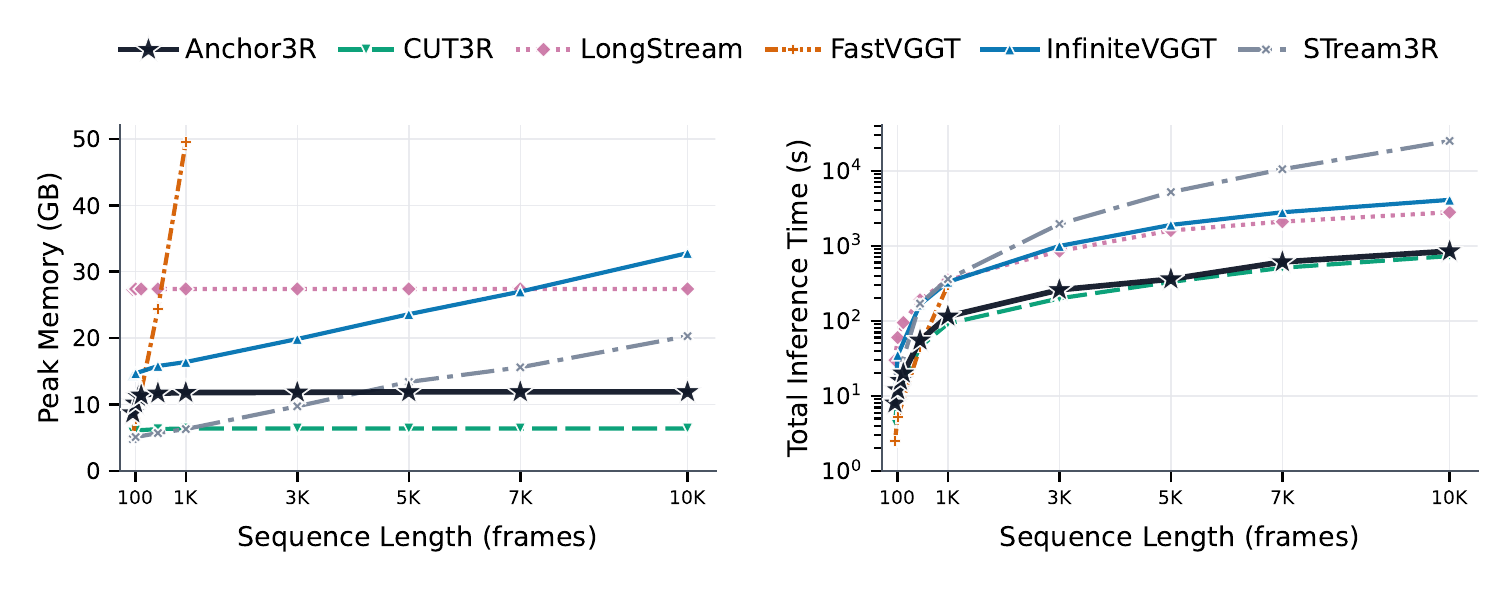}
    \caption{
    Runtime and GPU memory scaling under the same resolution and hardware.
    Anchor3R, trained on 48-frame sequences, maintains nearly constant GPU memory and approximately linear runtime up to 10K frames.
    }
    \label{fig:runtime_memory}
    \vspace{-0.2cm}
\end{figure}

\paragraph{Runtime and memory scaling.}
Figure~\ref{fig:runtime_memory} compares Anchor3R with CUT3R~\citep{wang2025continuous}, LongStream~\citep{cheng2026longstream}, FastVGGT~\citep{shen2025fastvggt}, InfiniteVGGT~\citep{yuan2026infinitevggt}, and STream3R~\citep{lan2025stream3r} under the same resolution and hardware. While several baselines exhibit increasing memory consumption or substantially higher runtime as the sequence grows, Anchor3R maintains nearly constant GPU memory up to 10K frames and scales approximately linearly in runtime. Notably, the model is trained on 48-frame sequences but directly generalizes to streams of up to 10K frames in this evaluation. The full model runs at 12.7 FPS with 11.9\,GB peak GPU memory. Its runtime is comparable to CUT3R, while its GPU memory remains substantially lower than LongStream and InfiniteVGGT at long horizons. The accumulated motion graph and pointmaps are stored on CPU/disk and grow linearly with sequence length. Offline motion averaging remains lightweight, taking 11.88\,s for 12K frames and 17.24\,s for 18.8K frames.

\paragraph{Controlled offline refinement.}
To isolate the effect of offline refinement, we apply the same loop-keyframe interval ($k=5$) and motion-averaging protocol to LongStream and VGGT. LongStream-Off-5 improves over LongStream on 6/7 VBR sequences, while VGGT-Off-5 improves over VGGT-Long on all seven, confirming that loop-aware refinement benefits different prediction models. Under the same protocol, Anchor3R-Off-5 outperforms LongStream-Off-5 on all seven sequences and VGGT-Off-5 on 6/7 sequences, indicating that its offline gains cannot be attributed solely to the refinement backend. We further vary the keyframe interval $k\in\{5,10,15\}$ and observe consistent performance across most sequences, suggesting limited sensitivity to this choice.

\section{Limitations}

Anchor3R has several limitations. First, historical pointmaps are not updated after their initial prediction, which may limit global geometric consistency. Second, the current motion-averaging formulation assumes reasonably consistent relative scales and may degrade when scale estimates are unreliable. Third, although VBR, KITTI, and Waymo contain moving objects, we do not conduct dedicated evaluation on highly dynamic benchmarks, and robustness to severe scene dynamics remains to be studied. 
Finally, loop retrieval and global pose alignment remain separate from the learned reconstruction model; jointly learning long-range constraint discovery and global alignment is a promising direction toward a more fully end-to-end system.
Our evaluation is also replay-based and does not include deployment on a physical robot.

\section{Conclusion}

We presented \emph{Anchor3R}, a current-centric streaming 3D reconstruction framework for long-horizon visual mapping. Anchor3R treats feed-forward reconstruction as local measurement prediction: it predicts window-relative poses anchored at the current frame, caches reusable image-conditioned states while re-instantiating pose queries for each window, and integrates the resulting measurements through online pose updates and loop-aware motion averaging. This design avoids persistent global-gauge propagation inside the network, reduces stale pose-cache interference, and enables global drift redistribution through an explicit motion graph. Experiments on indoor, outdoor, driving, and RGB-D benchmarks show that
Anchor3R improves long-horizon pose accuracy and dense map alignment over existing streaming feed-forward baselines, 
while controlled analyses validate the benefits of current-centric prediction, image-only caching, and graph-based pose recovery, as well as the scalability of Anchor3R to long visual streams.


\acknowledgments{
This work was supported by the National Natural Science Foundation of China
(Nos. U23A20386, 62572470, and U22B2055) and the Beijing Natural Science
Foundation (No. L223003).
}

\bibliography{example}

\clearpage
\appendix
\section*{Supplementary Material}

\section{Online Motion Averaging}
\label{app:online_motion_averaging}

During online inference, Anchor3R does not compose poses along a single temporal chain. 
Instead, the current pose is estimated from all available window-relative measurements to previously recovered frames. 
At time $t$, for each historical frame $i\in\mathcal{W}_t\setminus\{t\}$, the network predicts a relative rotation
$\hat{\mathbf{R}}_{i\leftarrow t}$ satisfying
\begin{equation}
\hat{\mathbf{R}}_{i\leftarrow t}\approx \mathbf{R}_i\mathbf{R}_t^\top .
\end{equation}
Given the previously estimated global rotation $\hat{\mathbf{R}}_i$, this edge induces a candidate current rotation
\begin{equation}
\tilde{\mathbf{R}}_t^{(i)}
=
\hat{\mathbf{R}}_{i\leftarrow t}^{\top}\hat{\mathbf{R}}_i .
\end{equation}
Rather than selecting one candidate or averaging rotations in Euclidean space, we compute a robust Lie-algebra median over the candidate rotations
$\{\tilde{\mathbf{R}}_t^{(i)}\}_{i\in\mathcal{W}_t\setminus\{t\}}$.
This makes the online update robust to occasional inaccurate relative-pose predictions caused by weak overlap, motion blur, or transient matching ambiguity.

Specifically, we initialize multiple rotation hypotheses by randomly sampling from the candidate set. 
For each hypothesis $\mathbf{R}^{(m)}$, we compute residual rotations
$\tilde{\mathbf{R}}_t^{(i)}(\mathbf{R}^{(m)})^\top$ and map them to the Lie algebra:
\begin{equation}
\mathbf{r}_i^{(m)}
=
\operatorname{Log}\!\left(
\tilde{\mathbf{R}}_t^{(i)}(\mathbf{R}^{(m)})^\top
\right)\in\mathbb{R}^3 .
\end{equation}
The hypothesis is updated by the coordinate-wise median residual:
\begin{equation}
\Delta\boldsymbol{\omega}^{(m)}
=
\operatorname{median}_{i}\{\mathbf{r}_i^{(m)}\},
\qquad
\mathbf{R}^{(m)}
\leftarrow
\operatorname{Exp}(\Delta\boldsymbol{\omega}^{(m)})\mathbf{R}^{(m)} .
\end{equation}
The iteration stops when $\|\Delta\boldsymbol{\omega}^{(m)}\|<\epsilon$. 
Among all converged hypotheses, we select the one with the smallest median angular residual. 
We use stable implementations of the SO(3) logarithm and exponential maps, including small-angle handling, and use multiple random initializations to reduce sensitivity to the initial hypothesis. 
If no hypothesis converges, we fall back to the first candidate rotation.
We summarize this robust online rotation update in Algorithm~\ref{alg:online_rotation_median}.

The current camera center is estimated in the same robust manner but does not require iterative optimization. 
Each historical frame provides a candidate current center by adding its recovered position to the predicted relative translation:
\begin{equation}
\tilde{\mathbf{c}}_t^{(i)}
=
\hat{\mathbf{c}}_i+\hat{\mathbf{v}}_{i,t},
\qquad
\hat{\mathbf{v}}_{i,t}\approx \mathbf{c}_t-\mathbf{c}_i .
\end{equation}
We then take the coordinate-wise median over all candidates,
\begin{equation}
\hat{\mathbf{c}}_t
=
\operatorname{median}_{i\in\mathcal{W}_t\setminus\{t\}}
\{\tilde{\mathbf{c}}_t^{(i)}\}.
\end{equation}
This yields a simple and robust online translation update, consistent with the median-based rotation estimate. 
Together, the rotation and translation medians allow Anchor3R to integrate multiple current-centric measurements at each step without relying on a single fragile temporal edge.

\begin{algorithm}[t]
\caption{Online Lie-Algebra Median Rotation Averaging}
\label{alg:online_rotation_median}
\begin{algorithmic}[1]
\Require Candidate rotations $\{\tilde{\mathbf{R}}_j\}_{j=1}^{M}$, number of initializations $N_{\mathrm{init}}$, maximum iterations $K$, threshold $\epsilon$
\Ensure Robust current rotation $\hat{\mathbf{R}}_t$
\State Randomly initialize $\{\mathbf{R}^{(m)}\}_{m=1}^{N_{\mathrm{init}}}$ from $\{\tilde{\mathbf{R}}_j\}_{j=1}^{M}$
\State Set scores $e^{(m)}\leftarrow+\infty$ and convergence flags $c^{(m)}\leftarrow\mathrm{false}$
\For{$k=1$ to $K$}
    \For{$m=1$ to $N_{\mathrm{init}}$}
        \If{$c^{(m)}=\mathrm{true}$}
            \State \textbf{continue}
        \EndIf
        \For{$j=1$ to $M$}
            \State $\mathbf{r}_j^{(m)}\leftarrow
            \operatorname{Log}\!\left(\tilde{\mathbf{R}}_j(\mathbf{R}^{(m)})^\top\right)$
            \State $\theta_j^{(m)}\leftarrow \|\mathbf{r}_j^{(m)}\|_2$
        \EndFor
        \State $\Delta\boldsymbol{\omega}^{(m)}
        \leftarrow \operatorname{median}_{j}\{\mathbf{r}_j^{(m)}\}$
        \If{$\|\Delta\boldsymbol{\omega}^{(m)}\|_2 < \epsilon$}
            \State $e^{(m)}\leftarrow \operatorname{median}_{j}\{\theta_j^{(m)}\}$
            \State $c^{(m)}\leftarrow\mathrm{true}$
        \Else
            \State $\mathbf{R}^{(m)}\leftarrow
            \operatorname{Exp}(\Delta\boldsymbol{\omega}^{(m)})\mathbf{R}^{(m)}$
        \EndIf
    \EndFor
    \If{all hypotheses have converged}
        \State \textbf{break}
    \EndIf
\EndFor
\If{at least one hypothesis converged}
    \State $m^\star\leftarrow\arg\min_m e^{(m)}$
    \State $\hat{\mathbf{R}}_t\leftarrow \mathbf{R}^{(m^\star)}$
\Else
    \State $\hat{\mathbf{R}}_t\leftarrow \tilde{\mathbf{R}}_1$
\EndIf
\State \Return $\hat{\mathbf{R}}_t$
\end{algorithmic}
\end{algorithm}

\section{Offline Motion Averaging}
\label{app:Offline Motion Averaging}

After streaming inference, Anchor3R refines the complete trajectory by optimizing over the accumulated window-relative measurements.
Unlike online pose averaging, which estimates each new pose from only the current active window, offline motion averaging uses all local edges produced along the sequence, including additional long-range edges from loop-closure reinsertion.
This step is enabled by our current-centric task definition: each prediction is a reusable relative-pose constraint rather than an absolute pose tied to a historical coordinate system.

\paragraph{Rotation averaging.}
Let $F$ be the number of frames and let $S=F-W+1$ be the number of sliding windows.
For each window $s\in\{1,\dots,S\}$ and each source frame $j\in\{0,\dots,W-2\}$, Anchor3R predicts a relative rotation
$\hat{\mathbf{R}}_{s+j\leftarrow s+W-1}$, where the last frame of the window is the current anchor.
We initialize the global rotations with the online estimates and remove the gauge by setting the frame at index $W-1$ as the reference:
\begin{equation}
\mathbf{R}_i^{(0)}
=
\hat{\mathbf{R}}^{\mathrm{online}}_i
\left(\hat{\mathbf{R}}^{\mathrm{online}}_{W-1}\right)^\top .
\end{equation}
For each relative rotation edge, the residual is computed in the Lie algebra as
\begin{equation}
\mathbf{r}_{s,j}
=
\operatorname{Log}\!\left(
(\mathbf{R}_{s+j})^\top
\hat{\mathbf{R}}_{s+j\leftarrow s+W-1}
\mathbf{R}_{s+W-1}
\right)\in\mathbb{R}^3 .
\end{equation}
The gauge-fixing residual is
$\mathbf{r}_{\mathrm{fix}}=\operatorname{Log}(\mathbf{R}_{W-1}^{\top})$.
At each IRLS iteration, we solve for incremental axis-angle updates
$\{\Delta\boldsymbol{\omega}_i\}_{i=1}^{F}$ with a sparse linearized system.
For an edge $(s+j, s+W-1)$, the corresponding linearized constraint is
\begin{equation}
\Delta\boldsymbol{\omega}_{s+j}
-
\Delta\boldsymbol{\omega}_{s+W-1}
\approx
\mathbf{r}_{s,j}.
\end{equation}
We apply a robust Geman--McClure-style weight to each residual,
\begin{equation}
w_{s,j}
=
\frac{\sigma^2}{(\sigma^2+\|\mathbf{r}_{s,j}\|_2^2)^2},
\qquad
\sigma=5^\circ ,
\end{equation}
and solve the weighted normal equation
\begin{equation}
(\mathbf{A}^{\top}\mathbf{W}\mathbf{A})\Delta\boldsymbol{\omega}
=
\mathbf{A}^{\top}\mathbf{W}\mathbf{r}.
\end{equation}
The rotations are then updated by right multiplication:
\begin{equation}
\mathbf{R}_i
\leftarrow
\mathbf{R}_i\operatorname{Exp}(\Delta\boldsymbol{\omega}_i).
\end{equation}
We repeat this process until the average update magnitude is below a small threshold or the maximum number of iterations is reached.

\paragraph{Position averaging.}
After rotation averaging, we recover camera centers from all window-relative translation measurements. 
For each edge $(s+j, s+W-1)$, we first rotate the predicted relative translation into the global frame:
\begin{equation}
\hat{\mathbf{v}}_{s,j}
=
\mathbf{R}_{s+j}^{\top}
\hat{\mathbf{t}}_{s+j\leftarrow s+W-1}
\approx
\mathbf{c}_{s+W-1}-\mathbf{c}_{s+j}.
\end{equation}
We then estimate the global camera centers by directly enforcing the relative translation constraints:
\begin{equation}
\mathbf{c}_{s+j}
-
\mathbf{c}_{s+W-1}
+
\hat{\mathbf{v}}_{s,j}
=
\mathbf{0},
\qquad
j=0,\dots,W-2 .
\end{equation}
The trajectory gauge is fixed by setting one reference center, e.g., $\mathbf{c}_{W-1}=\mathbf{0}$.
This gives a sparse L1 translation averaging problem:
\begin{equation}
\min_{\{\mathbf{c}_i\}}
\sum_{s=1}^{S}\sum_{j=0}^{W-2}
\left\|
\mathbf{c}_{s+j}
-
\mathbf{c}_{s+W-1}
+
\hat{\mathbf{v}}_{s,j}
\right\|_1,
\qquad
\mathrm{s.t.}\quad \mathbf{c}_{W-1}=\mathbf{0}.
\end{equation}
Here the L1 objective improves robustness to inaccurate relative translations from weak-overlap or ambiguous windows.
Unlike the scale-variable formulation, this direct form does not allow each window to absorb errors with an independent scale, and therefore better preserves the scale consistency learned between relative poses and local pointmaps.
In implementation, we assemble the constraints into a sparse linear system and solve the LAD problem with ADMM, where the $x$-step is accelerated by sparse Cholesky factorization.
The resulting camera centers, together with the averaged rotations, define the final global trajectory used to align local pointmaps into a coherent reconstruction.
\begin{algorithm}[t]
\caption{Offline Motion Averaging}
\label{alg:offline_motion_averaging}
\begin{algorithmic}[1]
\Require Window-relative poses $\{\hat{\mathbf{R}}_{s+j\leftarrow s+W-1}, \hat{\mathbf{t}}_{s+j\leftarrow s+W-1}\}$, online rotations $\{\hat{\mathbf{R}}^{\mathrm{online}}_i\}$, window size $W$
\Ensure Refined global rotations $\{\mathbf{R}_i\}$ and camera centers $\{\mathbf{c}_i\}$
\State Initialize $\mathbf{R}_i \leftarrow \hat{\mathbf{R}}^{\mathrm{online}}_i(\hat{\mathbf{R}}^{\mathrm{online}}_{W-1})^\top$
\State Build sparse rotation incidence matrix $\mathbf{A}$ with gauge fixed at frame $W-1$
\For{$k=1$ to $K_{\mathrm{rot}}$}
    \State Compute Lie residuals
    $\mathbf{r}_{s,j}\leftarrow
    \operatorname{Log}((\mathbf{R}_{s+j})^\top
    \hat{\mathbf{R}}_{s+j\leftarrow s+W-1}
    \mathbf{R}_{s+W-1})$
    \State Compute robust weights
    $w_{s,j}\leftarrow \sigma^2/(\sigma^2+\|\mathbf{r}_{s,j}\|_2^2)^2$
    \State Solve
    $(\mathbf{A}^{\top}\mathbf{W}\mathbf{A})\Delta\boldsymbol{\omega}
    =
    \mathbf{A}^{\top}\mathbf{W}\mathbf{r}$
    \State Update $\mathbf{R}_i\leftarrow \mathbf{R}_i\operatorname{Exp}(\Delta\boldsymbol{\omega}_i)$ for all frames
    \If{$\frac{1}{F}\sum_i\|\Delta\boldsymbol{\omega}_i\|_2 < \epsilon_{\mathrm{rot}}$}
        \State \textbf{break}
    \EndIf
\EndFor
\State Rotate relative translations into the global frame:
$\hat{\mathbf{v}}_{s,j}\leftarrow \mathbf{R}_{s+j}^{\top}\hat{\mathbf{t}}_{s+j\leftarrow s+W-1}$
\State Build sparse LAD system over camera centers $\{\mathbf{c}_i\}$ with fixed scale, enforcing
$\mathbf{c}_{s+j}-\mathbf{c}_{s+W-1}+\hat{\mathbf{v}}_{s,j}\approx\mathbf{0}$
\State Solve
$\min_{\mathbf{c}}\|\mathbf{A}_{t}\mathbf{c}-\mathbf{b}_{t}\|_1$
with ADMM and sparse Cholesky
\State \Return $\{\mathbf{R}_i\},\{\mathbf{c}_i\}$
\end{algorithmic}
\end{algorithm}

\section{Additional Ablation on Online and Offline Motion Averaging}
\label{app:ablation_online_offline}

We further ablate the effect of offline motion averaging without loop closure on KITTI. 
This setting isolates whether the dense window-relative measurements accumulated during online streaming can improve pose consistency by themselves, without introducing additional long-range loop constraints. 

\begin{table*}[t]
\centering
\caption{
\textbf{Additional ablation on KITTI.}
We compare Anchor3R-Online and Anchor3R-Offline without loop closure on all 11 KITTI Odometry sequences.
Offline w/o LC optimizes only the dense window-relative measurements accumulated during streaming inference, without loop-closure reinsertion.
Best RRE and RTE values between online and offline variants are highlighted in bold.
}
\label{tab:kitti_online_offline_rre_rte}
\begin{tabular}{c c | cc | cc}
\toprule
\multirow{2}{*}{Sequence}
& \multirow{2}{*}{\#Frames}
& \multicolumn{2}{c|}{Pose-Online}
& \multicolumn{2}{c}{Pose-Offline w/o LC} \\
\cmidrule(lr){3-4}
\cmidrule(lr){5-6}
&
& RRE $(^\circ)\downarrow$
& RTE $(m)\downarrow$
& RRE $(^\circ)\downarrow$
& RTE $(m)\downarrow$ \\
\midrule
00 & 4541 & 0.200 & 0.084 & \textbf{0.191} & \textbf{0.072} \\
01 & 1101 & 0.220 & 0.334 & \textbf{0.203} & \textbf{0.305} \\
02 & 4661 & 0.198 & \textbf{0.234} & \textbf{0.188} & \textbf{0.234} \\
03 & 801  & 0.152 & 0.072 & \textbf{0.137} & \textbf{0.063} \\
04 & 271  & 0.129 & 0.137 & \textbf{0.120} & \textbf{0.116} \\
05 & 2761 & 0.156 & 0.105 & \textbf{0.149} & \textbf{0.096} \\
06 & 1101 & 0.141 & 0.105 & \textbf{0.134} & \textbf{0.087} \\
07 & 1101 & 0.174 & 0.082 & \textbf{0.166} & \textbf{0.075} \\
08 & 4071 & 0.182 & 0.148 & \textbf{0.168} & \textbf{0.144} \\
09 & 1591 & 0.212 & 0.231 & \textbf{0.202} & \textbf{0.229} \\
10 & 1201 & 0.180 & 0.081 & \textbf{0.169} & \textbf{0.070} \\
\midrule
Avg. & -- & 0.177 & 0.147 & \textbf{0.166} & \textbf{0.136} \\
\bottomrule
\end{tabular}
\end{table*}

As shown in Table~\ref{tab:kitti_online_offline_rre_rte}, offline motion averaging consistently improves pose accuracy even without loop closure.
Compared with the online trajectory, Offline w/o LC reduces the average RRE from $0.177^\circ$ to $0.166^\circ$ and the average RTE from $0.147$\,m to $0.136$\,m.
This confirms that the current-centric dense relative-pose predictions provide useful redundant constraints for global pose refinement, rather than serving only as one-step online estimates.

\section{Controlled Offline Refinement on VBR}
\label{app:controlled_offline}
\begin{table*}[t]
\centering
\caption{
\textbf{Controlled offline refinement on VBR.}
We apply the same loop-keyframe retrieval and motion-averaging protocol to
LongStream, VGGT, and Anchor3R. Off-$k$ denotes a loop-keyframe interval of $k$.
Lower ATE is better.
}
\label{tab:vbr_controlled_offline}
\resizebox{\textwidth}{!}{
\begin{tabular}{l|ccccccc|c}
\toprule
Method
& campus\_train0
& campus\_train1
& ciampino\_train1
& colosseo\_train0
& diag\_train0
& pincio\_train0
& spagna\_train0
& Avg. \\
\midrule
LongStream
& 100.57 & 105.55 & 131.78 & 72.52 & 32.35 & 43.47 & 59.31 & 77.93 \\
LongStream-Off-5
& 87.61 & 99.86 & 98.64 & 45.99 & 8.46 & 65.84 & 39.77 & 63.74 \\
\midrule
VGGT-Long
& 118.59 & 98.21 & 172.13 & 39.56 & 30.80 & 53.44 & 50.27 & 80.43 \\
VGGT-Off-5
& 12.20 & 9.85 & 28.30 & 20.21 & 12.84 & 19.77 & 15.56 & 16.96 \\
\midrule
Anchor3R-Off-5
& 5.13 & 3.52 & 78.64 & 17.25 & 5.35 & 15.56 & 11.76 & 19.60 \\
Anchor3R-Off-10
& 5.02 & 2.91 & 92.51 & 19.15 & 5.21 & 16.51 & 11.27 & 21.80 \\
Anchor3R-Off-15
& 5.15 & 3.62 & 100.87 & 21.65 & 5.04 & 18.77 & 12.11 & 23.89 \\
\bottomrule
\end{tabular}
}
\end{table*}

Applying the same offline refinement to other prediction models also improves
their trajectories: LongStream-Off-5 improves over LongStream on 6 of 7
sequences, while VGGT-Off-5 improves over VGGT-Long on all seven.
Under the same protocol, Anchor3R-Off-5 outperforms LongStream-Off-5 on all
seven sequences and VGGT-Off-5 on 6 of 7 sequences, indicating that the gains
cannot be attributed solely to the shared refinement backend.

We further vary the loop-keyframe interval for Anchor3R.
Performance remains broadly consistent on six of seven sequences for
$k\in\{5,10,15\}$, while \textit{ciampino\_train1} becomes more challenging
as the interval increases. This suggests that Anchor3R is generally insensitive
to moderate changes in loop-keyframe density, although difficult trajectories
can benefit from denser long-range constraints.

\section{Failure Case Analysis}
\label{app:failure_case}
\begin{figure}[t]
    \centering
    \includegraphics[width=\linewidth]{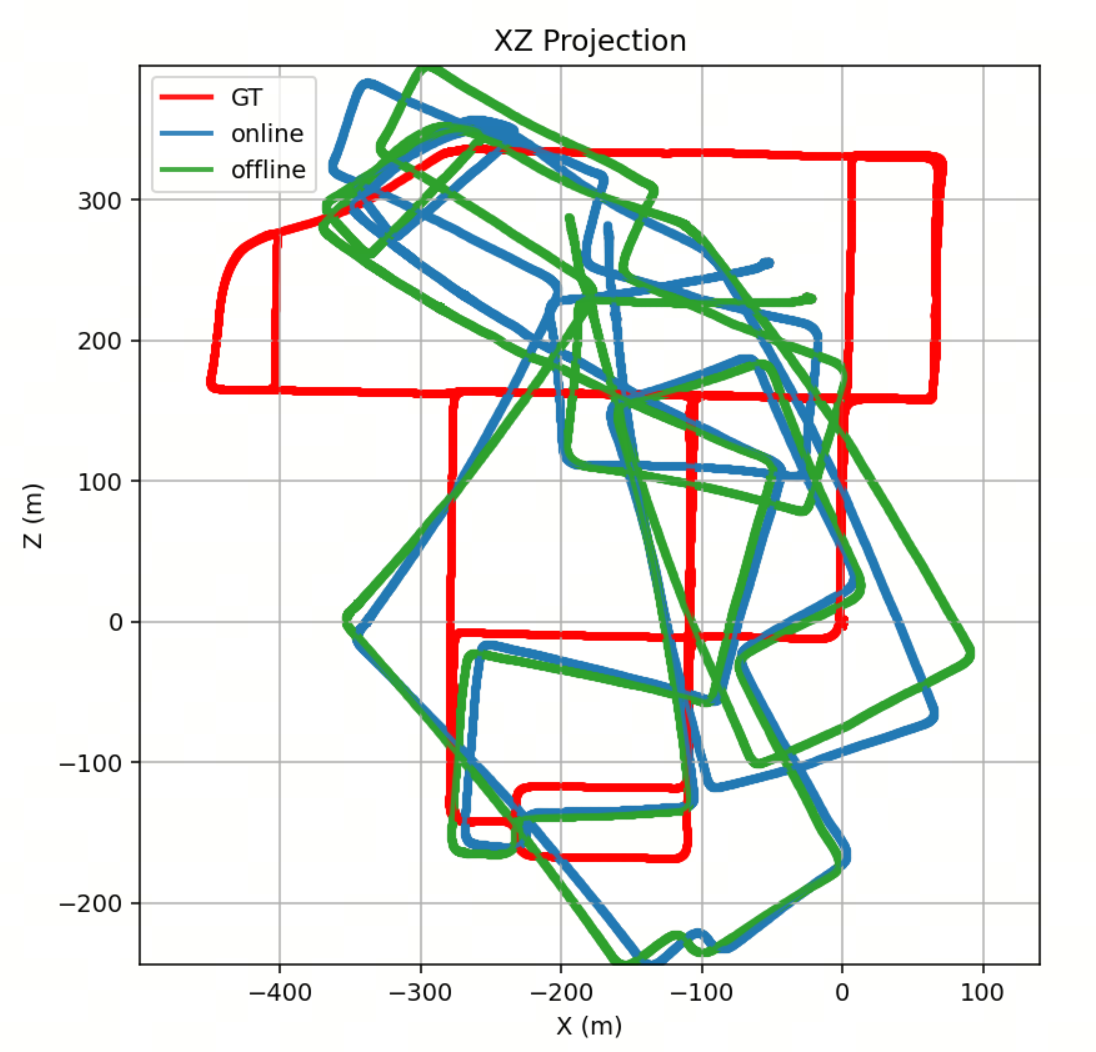}
    \caption{
    \textbf{Failure case on \textit{ciampino\_train1}.}
    We show the online trajectory and offline refinement without loop-closure
    constraints. Errors are most evident around frequent turns, where
    orientation errors affect subsequent translation directions and accumulate
    over the long trajectory.
    }
    \label{fig:failure_case}
\end{figure}
Figure~\ref{fig:failure_case} shows a representative failure case on
\textit{ciampino\_train1}. Both the online result and offline refinement without
loop closure exhibit noticeable drift around regions with frequent turns.
Small orientation errors alter the directions of subsequent translations and
can therefore accumulate over a long trajectory. Offline motion averaging
redistributes locally redundant measurements, but without additional long-range
constraints it cannot fully eliminate this accumulated drift. This example
highlights the remaining challenge of maintaining robust orientation estimates
under repeated large rotations and motivates stronger long-range constraint
discovery in future work.

\section{Evaluation Dataset Details}

All evaluations are conducted on complete sequences without frame subsampling. 
We summarize the dataset-specific settings below.

\textbf{VBR}. 
Following the protocol of LoGeR~\citep{zhang2026loger}, we use all 7 full-length sequences, which contain 8,815--18,846 frames and cover trajectories up to 5.2\,km.

\textbf{KITTI}. 
For KITTI Odometry, we evaluate the full camera-02 image streams from all 11 sequences, i.e., sequences 00--10.

\textbf{Waymo Open}. 
We evaluate on 9 held-out segments that are excluded from our training set: 163453191 (198 frames, 160 m), 183829460 (199 frames, 42 m), 315615587 (199 frames, 165 m), 346181117 (199 frames, 351 m), 371159869 (196 frames, 273 m), 405841035 (199 frames, 86 m), 460417311 (198 frames, 266 m), 520018670 (199 frames, 135 m), and 610454533 (198 frames, 63 m). 
Although Waymo is included in the training data, these held-out segments are used to test generalization to unseen driving scenes.

\textbf{Oxford Spires}. 
We use the front-camera images and evaluate on all 12 subsets.

\textbf{7Scenes}. 
For each scene in 7Scenes, including Chess, Fire, Heads, Office, Pumpkin, RedKitchen, and Stairs, we evaluate on sequence 01.

\textbf{TUM RGB-D}. 
We follow the standard full-sequence evaluation protocol.



\end{document}